\documentclass{article}

\usepackage[preprint]{corl_2025} 

\usepackage{lipsum}
\usepackage{wrapfig}
\usepackage{multirow}
\usepackage{booktabs}
\usepackage{rotating}

\usepackage{amsfonts}
\usepackage{amsmath}
\usepackage{wrapfig}
\usepackage{graphicx}
\usepackage{lipsum} 
\usepackage{wrapfig,booktabs}

\usepackage{bbm}

\title{\LARGE \bf
DreamGrasp:\\ Zero-Shot 3D Multi-Object Reconstruction from Partial-View Images for Robotic Manipulation}

\author{
  Young Hun Kim$^{1}$ \:\: Seungyeon Kim$^{1}$ \:\: Yonghyeon Lee$^{2}$ \:\: Frank Chongwoo Park$^{1}$\\
  $^{1}$Seoul National University, $^{2}$Massachusetts Institute of Technology\\
  \texttt{\{yhun, ksy\}@robotics.snu.ac.kr, yhl@mit.edu, fcp@snu.ac.kr}
}

\begin{document}

\maketitle

\begin{abstract}
Partial-view 3D recognition -- reconstructing 3D geometry and identifying object instances from a few sparse RGB images -- is an exceptionally challenging yet practically essential task, particularly in cluttered, occluded real-world settings where full-view or reliable depth data are often unavailable. 
Existing methods, whether based on strong symmetry priors or supervised learning on curated datasets, fail to generalize to such scenarios. 
In this work, we introduce DreamGrasp, a framework that leverages the imagination capability of large-scale pre-trained image generative models to infer the unobserved parts of a scene. 
By combining coarse 3D reconstruction, instance segmentation via contrastive learning, and text-guided instance-wise refinement, DreamGrasp circumvents limitations of prior methods and enables robust 3D reconstruction  in complex, multi-object environments. Our experiments show that DreamGrasp not only recovers accurate object geometry but also supports downstream tasks like sequential decluttering and target retrieval with high success rates.
\end{abstract}


\section{INTRODUCTION}

Robot tasks such as target-driven manipulation and 3D spatial reasoning -- like collision avoidance and path planning -- require the ability to reconstruct the 3D geometry of objects and identify object instances from visual observations.
While many existing methods rely on accurate depth~\cite{kim2022dsqnet, sajjan2020clear} or all-around-view RGB images~\cite{ichnowski2021dex, kerr2022evo, lee2023nfl}, we address a more practical setting: reconstruction from only sparse, partial-view RGB images (e.g., as few as two).
This direction is crucial for real-world deployment, where depth sensing often fails for transparent or reflective objects, and full 360° scene coverage is rarely available due to occlusions in cluttered environments or the limited workspace of a robot.

Recent approaches address this challenge by employing supervised learning on datasets of partial observations paired with complete scenes~\cite{kim2024t2sqnet, dai2023graspnerf}. However, the performance of these supervised learning approaches is inherently constrained by the diversity of the datasets.
This highlights the need for alternative approaches that can generalize more robustly beyond the constraints of curated 3D training data.

We draw inspiration from recent findings that large-scale image generative models -- when suitably fine-tuned with datasets of paired camera poses and images -- can produce reasonable predictions of unseen parts of a scene (e.g., the backs of objects) given a partial-view image as input~\cite{liu2023zero, qian2023magic123, tang2023dreamgaussian, liu2023one, liu2023syncdreamer, long2024wonder3d, tang2024mvdiffusion++}.
However, prior works demonstrate strong performance only on clean, front-view images of single, unoccluded objects, and often struggle in multi-object scenarios where objects heavily occlude one another -- which are prevalent in robotic object manipulation tasks.

Our key idea is to first reconstruct a coarse 3D geometry from partial-view images, and then segment multiple objects into individual instances -- where we adopt the recent contrastive lifting algorithm~\cite{silva2024contrastive}, augmented with a novel surface regularization term -- followed by object-wise 3D geometry refinement.
Specifically, during the geometry refinement stage, we iteratively fit our model to the given images using an instance-wise RGB rendering loss, while simultaneously performing outlier removal. Most importantly, we incorporate additional biases into each object's geometry using text descriptions automatically generated by a Multimodal Large Language Model (MLLM) from the segmented images, guiding the reconstructed geometry to better align with the texts.
We refer to our framework as {\it DreamGrasp} -- a zero-shot 3D reconstruction and instance identification method from sparse, partial-view images, leveraging large-scale generative models without fine-tuning.

Through experiments, we demonstrate that DreamGrasp achieves reliable 3D shape recognition performance. Furthermore, by applying DreamGrasp to object manipulation tasks that require instance-wise geometric understanding of multiple objects in a scene -- such as sequential decluttering and target retrieval -- we show that the resulting 3D representations provide sufficiently accurate geometry for the robot to perform grasping and collision avoidance.

\begin{figure}[!t]
    \centering
    \vspace{-20pt}
    \includegraphics[width=\linewidth]{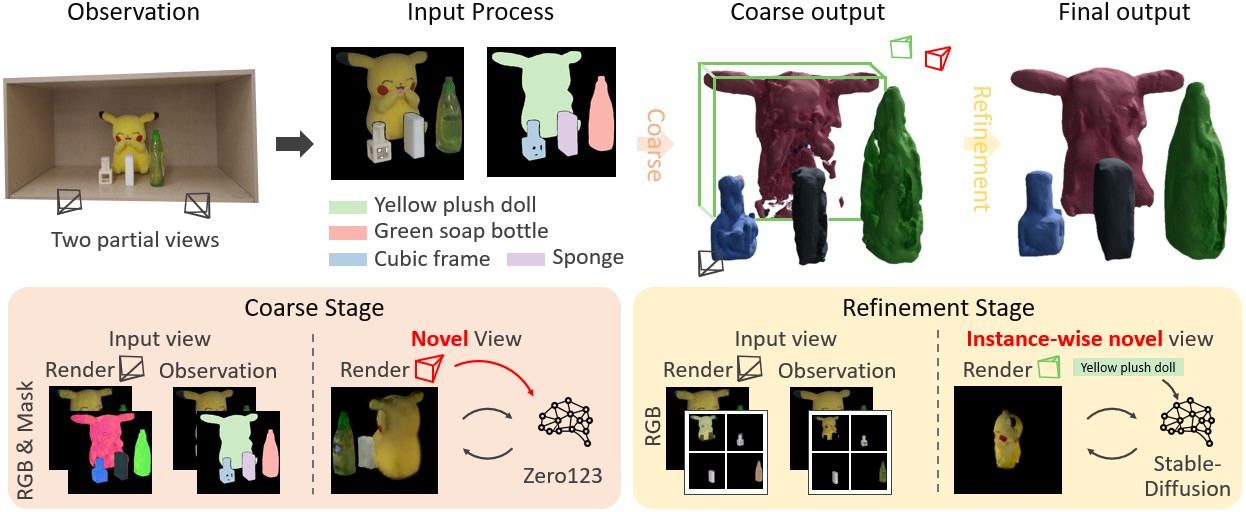}
    \vspace{-15pt}
    \caption{Overall pipeline of DreamGrasp. 
    ({\it Observation}) Our method uses only two partial-view RGB images as input.
    ({\it Input process}) Instance masks and text prompts are extracted from the RGB images using SAM~\cite{kirillov2023segany} and ChatGPT~\cite{openai2024gpt4o}, respectively.
    ({\it Coarse stage}) These inputs are used for initial scene-level geometry reconstruction, leveraging RGB and instance mask images with novel-view supervision guided by Zero123.
    ({\it Refinement stage}) The coarse scene is segmented using learned features, and each object is refined through instance-wise RGB input and novel-view supervision guided by a text-conditioned diffusion model.    
    }
    \vspace{-10pt}
    \label{fig:pipeline}
\end{figure}

\section{Preliminaries: Distilling Image Generative Models for 3D Reconstruction}
This section provides a brief overview of score distillation in our context -- a framework that transfers the knowledge of pre-trained 2D image diffusion models to the task of 3D model generation or reconstruction -- laying the foundation for our method.
We begin by describing how to fit a 3D model using all-around multi-view RGB images~\cite{mildenhall2021nerf, kerbl20233d}, then explain how score distillation enables this process with only sparse views~\cite{liu2023zero, qian2023magic123, tang2023dreamgaussian, liu2023one, liu2023syncdreamer, long2024wonder3d, tang2024mvdiffusion++}, and finally discuss its current limitations.

Let $\phi$ denote a 3D scene representation, such as an implicit field model that outputs color and occupancy at a query point -- e.g., Neural Radiance Fields (NeRF)~\cite{mildenhall2021nerf} -- or an explicit parametrization like a Gaussian mixture, as used in Gaussian Splatting~\cite{kerbl20233d}. We assume that $\phi$ can be rendered in a differentiable manner from a given camera pose $X$ to produce an image $I$, denoted by $\pi_X(\phi) = I$, where $\pi_X$ is differentiable.  
Given a set of images equipped with camera poses $\{(I_i, X_i)\}_{i=1}^{N}$, the model $\phi$ can be optimized by minimizing the average reconstruction loss:  
\begin{equation}
\label{eq:rendering_loss}
    \frac{1}{N} \sum_{i=1}^{N} \left\| \pi_{X_i}(\phi) - I_i \right\|^2.
\end{equation}
With sufficient multi-view coverage, high-quality $\phi$ can be recovered. However, with only sparse views, the optimization in Equation~\eqref{eq:rendering_loss} often fails to recover the true scene unless a strong prior is imposed on $\phi$, which is rarely practical due to the model’s flexibility and tendency to overfit limited observations.

To address this, Zero123~\cite{liu2023zero} leverages image-conditioned diffusion models, demonstrating the ability to generate plausible 3D models even from a single front-view image. 
In diffusion-based generative models, the image distribution is modeled via a score function $\nabla \log p(t, I_t)$, where $t \in \{1, \ldots, T\}$ and $I_t$ is the image at step $t$. Generation begins with a noisy sample $I_T$ and iteratively refines it toward $I_1$ by following the score direction with added noise. The score function provides local guidance for transforming the image to better match the underlying data distribution of a large image dataset.


Zero123 fine-tunes a large-scale generative model, Stable Diffusion~\cite{rombach2021highresolution}, to learn a conditional score function $\nabla \log p(t, I_t^X \mid I_{\mathrm{front}}, X)$, where $I_{\mathrm{front}}$ is a front-view image, $X$ is the relative camera pose, and $I_t^X$ is the image at diffusion step $t$ from viewpoint $X$. Using paired training data of images and poses, the model learns to guide noisy samples $I_t^X$ toward realistic views consistent with $I_{\mathrm{front}}$. This enables updating the 3D model $\phi$ not only to minimize the reconstruction loss in Equation~\eqref{eq:rendering_loss}, but also to follow the gradient of the conditional log-likelihood: 
\begin{equation} 
\label{eq:zero123} 
\nabla_{\phi} \log p(t, \pi_{X_i}(\phi)_t \mid I_{\mathrm{front}}, X_i), 
\end{equation} 
where $\pi_{X_i}(\phi)_t$ denotes the rendering of $\phi$ from pose $X_i$ at diffusion step $t$ (more precisely, with some approximation for computational efficiency; see~\cite{poole2022dreamfusion} for details on score distillation). 

However, the performance of Zero123 and its follow-up works~\cite{qian2023magic123, liu2023one, long2024wonder3d, shi2023zero123++} remains limited to single-object settings with clean input images, and often degrades in complex, multi-object scenes. This limitation stems from their reliance on fine-tuning datasets that are predominantly composed of single-object scenarios. Simply increasing dataset diversity for fine-tuning is challenging, as the diversity grows exponentially with the number of objects, making data collection prohibitively costly. This motivates us to introduce novel techniques to address this limitation while still leveraging the fine-tuned Zero123 model.

\section{DreamGrasp}

In this section, we present our \textit{DreamGrasp} framework. We begin by outlining the problem setting and key assumptions, followed by a detailed description of the overall pipeline and its components. The objective of DreamGrasp is to recover the 3D geometries of objects in a workspace and identify individual object instances from sparse, partial-view RGB images. We assume that the intrinsic and extrinsic parameters of the cameras are known. 

Figure~\ref{fig:pipeline} illustrates the overall DreamGrasp pipeline, which consists of (i) a coarse scene reconstruction stage and (ii) an instance-wise geometry refinement stage.
In the first stage, we adopt Zero123 to fit a coarse 3D scene geometry from a few sparse RGB images. Leveraging instance masks obtained from segmentation models such as SAM~\cite{kirillov2023segany}, we train an instance feature to enable 3D clustering of individual objects~\cite{silva2024contrastive}. As shown in the top-right of the figure, the resulting reconstruction is often noisy and geometrically inaccurate.
In the refinement stage, we crop object regions based on the instance feature and refine each object’s geometry independently. Each geometry is optimized using an instance-wise RGB rendering loss, along with score distillation from Zero123 and additional guidance from a Multimodal Large Language Model (MLLM) -- which generates text descriptions (upper left) -- via score distillation from a text-conditioned image generation model.

In the following sections, we detail each stage of the pipeline, including our novel Surface-Invariant Feature Regularizer and the instance-wise RGB reconstruction loss.

\subsection{Coarse Scene Reconstruction Stage}
We adopt Gaussian Splatting~\cite{kerbl20233d} to represent the 3D scene, augmented with an additional normalized $D_f$-dimensional feature channel to encode instance-aware features. We denote the set of 3D Gaussians as $\{ \mathcal{G}_i \}_{i=1}^{N_\mathcal{G}}$, where each Gaussian $\mathcal{G}_i = (\mu_i, \Sigma_i, \sigma_i, c_i, f_i)$ consists of a mean $\mu_i$, covariance matrix $\Sigma_i$, opacity $\sigma_i$, color $c_i$, and feature vector $f_i$. For simplicity, we refer to the entire set as $\mathcal{G}$. 
This representation is particularly advantageous for our purposes, as it enables effective 3D clustering via the learned feature vectors, allows a closed-form expression for our proposed regularizer, and facilitates instance-wise RGB rendering loss in the refinement stage.

In addition to the RGB rendering loss on the sparse input images, we leverage score distillation from Zero123~\cite{liu2023zero} to improve shape estimates of unseen regions, and apply contrastive learning~\cite{silva2024contrastive} using 2D instance masks to learn the 3D instance-aware features. In particular, the contrastive loss encourages Gaussians within the same mask to share similar feature values, while pushing apart those from different masks. This enforces feature consistency within each object instance and separability across objects. Further details are provided in Appendix B.

However, we find that the contrastive loss computed from sparse, partial-view images is insufficient to learn a smooth and accurate instance feature field, often leading to overfitting to the limited viewpoints. As a result, in some cases, feature vectors within the same instance become more dissimilar than those of spatially distant, unrelated objects, leading to incorrect 3D clustering results. To this end, we propose a novel regularization term, \textit{Surface-Invariant Feature Regularizer (SIFR)}, that encourages the Gaussian feature values to be as constant as possible along the object's surface. 

Specifically, we define the normalized feature field $\mathcal{F} : \mathbb{R}^3 \rightarrow S^{D_f-1}$ as
\begin{equation}
\mathcal{F}(x) = \frac{\mathcal{F}^*(x)}{\lVert \mathcal{F}^*(x)\rVert} 
\quad 
{\mathrm{where}} \quad
\mathcal{F}^*(x) = \sum_{i=1}^{N_\mathcal{G}} \exp\left(\frac{1}{2}(x-\mu_i)^T\Sigma_i^{-1}(x-\mu_i)\right)\sigma_i f_i.
\end{equation}
We then define SIFR as follows:
\begin{equation}
    \mathcal{L}_\text{SIFR} = \sum_{i=1}^{N_\mathcal{G}}\text{Tr}\left(\frac{\partial \mathcal{F}}{\partial x} (\mu_i)^T\,\frac{\partial \mathcal{F}}{\partial x} (\mu_i) \: G^{-1}(\mu_i)\right),
\end{equation}
which measures the local smoothness of $\mathcal{F}$ weighted by the input-space metric $G(x) \in \mathbb{R}^{3 \times 3}$, a positive-definite matrix. SIFR is more sensitive to variations in $\mathcal{F}$ along directions associated with smaller eigenvalues of $G(x)$. 

We construct the metric $G(x)$ as $G(x) = R(x) \Lambda R(x)^T$, where $\Lambda = \text{diag}(d, 1, 1)$ with $d \gg 1$ (e.g., $d=100$), and $R(x) \in \mathbb{R}^{3 \times 3}$ is a local rotation matrix. To encourage $\mathcal{F}(x)$ to remain invariant along the surface tangent directions, we set the first column of $R(x)$ to be the surface normal direction at $x$.
The surface normal can be estimated as $\nabla_x\rho / \lVert\nabla_x\rho\rVert$ where $\rho(\mathbf{x}) = \sum_{i=1}^{N_\mathcal{G}} \exp\left(\frac{1}{2}(x-\mu_i)^T\Sigma_i^{-1}(x-\mu_i)\right)\sigma_i$ is a density field.
The remaining two columns of $R(x)$ are any orthonormal vectors perpendicular to the surface normal. All terms admit closed-form expressions and can be efficiently computed.

\subsection{Instance-wise Refining Stage}

\begin{wrapfigure}{r}{0.5\textwidth}
    \centering
    \vspace{-35pt}
    \includegraphics[width=1\linewidth]{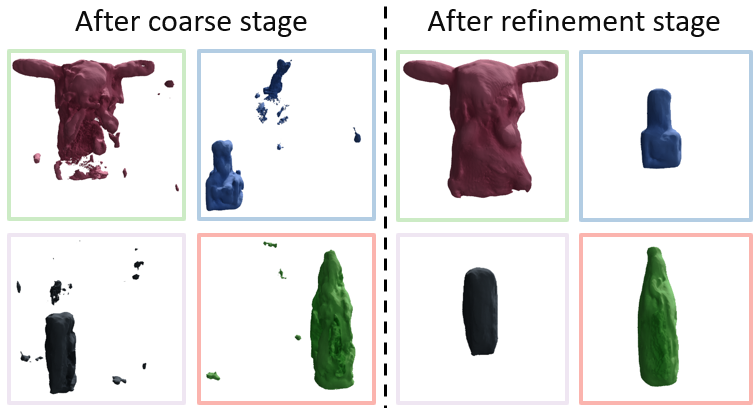}
    \caption{\textit{Left} : Segmentation results after the coarse stage, \textit{Right} :  Results of refinement stage.}
    \vspace{-10pt}
    \label{fig:segmentation_results}
\end{wrapfigure}

In the first stage, we have obtained the set of Gaussians $\{ \mathcal{G}_i = (\mu_i, \Sigma_i, \sigma_i, c_i, f_i) \}$, and now proceed to the instance-wise refinement stage.
We first perform 3D instance segmentation by clustering the Gaussians based on their learned feature vectors $f_i$. Specifically, we select a representative mask image at a given camera pose and render the Gaussian features to this view. For each $k$-th object, we compute the normalized mean feature vector $\overline{f}_k$ by averaging the features over the pixels corresponding to that object and then normalizing it.
We then define the Gaussians associated with the $k$-th object as those whose feature vectors exhibit high cosine similarity with $\overline{f}_k$: $
\mathcal{G}^k = \left\{ \mathcal{G}_i \in \mathcal{G} \;\middle|\; f_i \cdot \overline{f}_k \geq \delta \right\},$ where we use $\delta = 0.9$ in our experiments. However, as shown on the left of Figure~\ref{fig:segmentation_results}, the segmented Gaussians from the coarse stage are often noisy due to occlusions between objects. While our SIFR helps mitigate this issue to some extent, it remains insufficient, motivating the instance-wise refinement stage to further improve both segmentation and geometry quality.

We propose three components in the refinement stage. 
First, we introduce an instance-wise RGB reconstruction loss that aligns each object’s rendering with the observed RGB image under the corresponding mask. To enable this, we resolve instance correspondences across views by solving a bipartite matching problem based on mean rendered features, ensuring consistent object indexing across images (details in Appendix~C). The loss is defined as:
\begin{equation}
    \mathcal{L}_{\text{instance\_render}} = \sum_{j=1}^{N_I} \sum_{k=1}^{N_s} \left\| \pi_{X_j}(\mathcal{G}^k) \circ M_j^k - I_j \circ M_j^k \right\|^2,
\end{equation}
where $N_I$ is the number of observed images, $N_s$ is the number of object instances, and $\circ$ denotes elementwise multiplication between two images. Here, $M_j^k \in \{0, 1\}^{H \times W}$ is the $k$-th channel of the instance mask $M_j$, where a value of $1$ indicates that the $k$-th object occupies the pixel, and $0$ otherwise.

Second, based on our observation that applying Zero123 to each masked object image is insufficient when the mask is incomplete due to heavy occlusions by other objects, we introduce an additional score distillation strategy that provides a geometric bias to guide the refinement toward more complete and accurate object shapes.  
Specifically, we generate text descriptions for each object using a Multimodal Large Language Model (MLLM), and then apply score distillation using a text-conditioned diffusion model~\cite{poole2022dreamfusion}, aligning each instance's rendered appearance with its corresponding text prompt.

Lastly, we find that while score distillation naturally suppresses irrelevant Gaussians that misalign with the text and helps enhance incomplete geometry, some noisy Gaussians collapse into small, low-impact clusters that contribute little to the rendered image and are not effectively removed by distillation alone. To eliminate such outliers, we periodically apply a pointcloud-based outlier removal algorithm to the Gaussian centers for each instance. These refinement components are used in conjunction with novel view supervision from Zero123, which remains part of the optimization throughout the refinement stage.

\section{Object Manipulation with DreamGrasp}
In this section, to further highlight the strength of \textit{DreamGrasp} in recovering instance-wise geometry from partial views, we showcase two object manipulation tasks in a shelf environment: sequential decluttering and target retrieval.
We first present a method for sampling collision-free grasp poses from the instance-wise Gaussians $\mathcal{G}^k$ reconstructed by DreamGrasp, and then demonstrate how this grasp sampler and geometric information can be effectively used in downstream manipulation tasks.

Since we can render depth images of $\mathcal{G}^k$ from all-around views, we can apply TSDF fusion~\cite{zeng20163dmatch} with these depth images, resulting in various 3D representations, such as meshes, occupancy voxels, and point clouds. This compatibility enables the use of existing grasp planners~\cite{mahler2017dex, breyer2021volumetric, lim2024equigraspflow} developed for different types of 3D data to generate grasp poses for each object instance. In this work, we render depth images of $\mathcal{G}^k$ from all-around views and feed them into the pretrained depth-based grasp planner FC-GQ-CNN~\cite{satish2019policy} to generate diverse grasp poses. Among the sampled grasp poses, we retain only those that are collision-free throughout the approach trajectory, verified by checking for potential collisions with non-target objects. Further details on grasp pose sampling and collision checking are provided in Appendix D.

\textbf{Sequential Declutter} is the task of sequentially removing all objects in a scene through grasping.  
For reconstruction methods that do not provide instance-wise segmentation, it is often necessary to re-run the reconstruction process after each grasp, requiring reconstruction to be performed as many times as there are objects in the scene.  
In contrast, thanks to the instance-wise geometry provided by DreamGrasp, a single reconstruction suffices. This enables the entire decluttering sequence to be planned upfront from a single initial recognition step.

\textbf{Target Retrieval} involves extracting a specific target object that is initially ungraspable due to occlusions by surrounding objects. This requires rearranging obstacles within the workspace to make the target graspable. Prior works~\cite{kim2023leveraging, kim2024t2sqnet} have proposed solving this task using instance-wise 3D recognition, which requires an instance-wise grasp pose sampler, a grasp pose collision detector, and a model of rearrangement dynamics (i.e., predicting how the scene changes after rearrangement). The first two components are provided by DreamGrasp. For rearrangement, we adopt a simple pick-and-place formulation, where dynamics are approximated by moving the selected object from its current pose to a predefined placement pose. Such rearrangement dynamics would not be accessible without instance-level identification of objects.

\section{Experiments}

This experimental section serves two main purposes. First, we assess the 3D geometry estimation performance of DreamGrasp through both quantitative and qualitative evaluations in a controlled recognition setup. Second, we validate the effectiveness of DreamGrasp’s recognition results for real-world object manipulation. All experiments are conducted using only two partial-view RGB images as input for both DreamGrasp and the baselines.

\textbf{Baselines.}
We evaluate DreamGrasp against two baseline configurations, Contrastive Gaussian Clustering (CGC)~\cite{silva2024contrastive} and Zero123*, each corresponding to a subset of DreamGrasp's components. As such, this experiment functions as an ablation study to analyze the contribution of each component. \textit{CGC} proposes a method for jointly fitting a 3D scene with Gaussian Splatting and learning mask-based features through contrastive learning, given all-around-view RGB images and corresponding instance masks. We adapt CGC to the partial-view setting by providing only two front-view RGB images and corresponding instance masks, allowing us to evaluate the performance of vanilla Gaussian Splatting and contrastive learning under sparse observations. Also, 
we implement a baseline that combines Zero123-based 3D shape optimization from two partial views with the contrastive learning method from CGC. We denote this baseline as \textit{Zero123*} for convenience. For all baselines, 3D instance segmentation is performed using the same strategy as DreamGrasp, based on features learned through contrastive learning. 

\begin{figure}[!t]
    \centering
    \vspace{-20pt}
    \includegraphics[width=\linewidth]{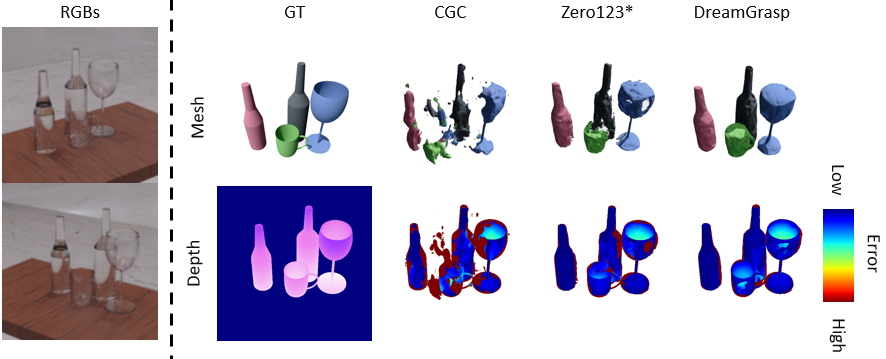}
    \vspace{-15pt}
    \caption{Recognition results from two partial RGB images from transparent TablewareNet objects. \textit{Upper}: Mesh from TSDF fusion with rendered depth images. \textit{Lower}: Ground truth depth image and depth error contours.}
    \vspace{-20pt}
    \label{fig:recog_results}
\end{figure}

\subsection{Recognition Performance}
\textbf{Datasets.}
To evaluate instance-wise recognition in cluttered scenes containing both opaque and transparent objects, we use object meshes from the YCB dataset~\cite{calli2015benchmarking} and the TablewareNet dataset~\cite{kim2024t2sqnet}. We simulate cluttered scenes by randomly placing 2 to 5 objects within a fixed-size workspace in the PyBullet simulator~\cite{coumans2016pybullet} and allowing them to settle. We then render two RGB images from front-view camera poses using Blender~\cite{blender}. We generate 20 scenes for each object count (2 to 5 objects), resulting in a total of 80 scenes.

\textbf{Instance Masks and Text Prompts.}
Since the experiments use synthetic data, we directly use simulation-derived ground-truth masks as input. Text prompts are assigned using mesh filenames for YCB and class labels for TablewareNet objects (e.g., “cup”, “wineglass”, “bottle”).

\textbf{Metrics.}
We evaluate the following three metrics:
\textbf{Scene Depth Accuracy} measures the accuracy of depth predictions from all-around novel views of the reconstructed scene. Here, depth accuracy is defined as the proportion of pixels where the predicted depth falls within a certain error threshold of the ground-truth depth. Accuracy is computed only on pixels where either the ground-truth or predicted object occupies space, excluding empty background pixels that dominate the image.
\textbf{Instance-wise Depth Accuracy} measures depth accuracy for each predicted instance over a scene where only that instance is present.
\textbf{Mask IoU} measures the IoU between the rendered mask from the predicted 3D scene and the ground-truth instance mask in the novel views. 
\begin{table*}[!h]
\vspace{-10pt}
\centering
\caption{Depth accuracy on scene-level and instance-level and 2D mask IoUs in cluttered scenes.}
\label{table:recognition_evaluation}
\begin{tabular}{lcccccccccc}

\multicolumn{2}{c}{}
&\multicolumn{3}{c}{Depth Acc. (Scene)}
&\multicolumn{1}{c}{}
&\multicolumn{3}{c}{Depth Acc. (Instance)}
&\multicolumn{2}{c}{}\\
\cline{3-5}
\cline{7-9}
\\
\multicolumn{1}{c}{\bf METHOD}
&\multicolumn{1}{c}{}
&\multicolumn{1}{c}{$\delta_{0.05}$}
&\multicolumn{1}{c}{$\delta_{0.10}$}
&\multicolumn{1}{c}{$\delta_{0.20}$}
&\multicolumn{1}{c}{}
&\multicolumn{1}{c}{$\delta_{0.05}$}
&\multicolumn{1}{c}{$\delta_{0.10}$}
&\multicolumn{1}{c}{$\delta_{0.20}$}
&\multicolumn{1}{c}{}
&\multicolumn{1}{c}{Mask IOU}
\\
\hline 
CGC && 0.527 & 0.575 & 0.586 && 0.467 & 0.493 & 0.495 && 0.482 \\
Zero123* && 0.747 & 0.789 & 0.793 && 0.666 & 0.694 & 0.695 && 0.705\\
DreamGrasp && \textbf{0.780} & \textbf{0.827} & \textbf{0.831} && \textbf{0.749} & \textbf{0.780} & \textbf{0.781} && \textbf{0.759} \\
\hline 
\end{tabular}
\vspace{-10pt}
\end{table*}

\textbf{Results.}
Table~\ref{table:recognition_evaluation} presents the quantitative recognition results, and Figure~\ref{fig:recog_results} provides a visual comparison of the outputs from each baseline. DreamGrasp consistently outperforms all baselines across varying object counts. While CGC struggles with geometry prediction under partial views, Zero123* performs competitively at the scene level. However, its performance drops significantly in instance-wise geometry prediction, emphasizing the importance of the refinement stage. Additional experimental details and extended results are provided in Appendix E.

\begin{figure}[!t]
    \centering
    \vspace{-20pt}
    \includegraphics[width=\linewidth]{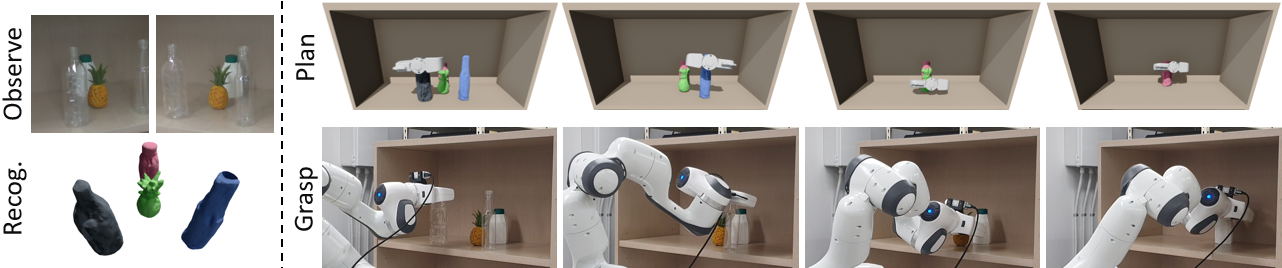}
    \vspace{-15pt}
    \caption{Example of real-world sequential decluttering on a shelf using single-shot recognition.}
    \vspace{-10pt}
    \label{fig:clear_clutter}
\end{figure}

\subsection{Object Manipulation Performance}
\begin{wrapfigure}{r}{0.3\textwidth}
    \centering
    \vspace{-55pt}
    \includegraphics[width=1\linewidth]{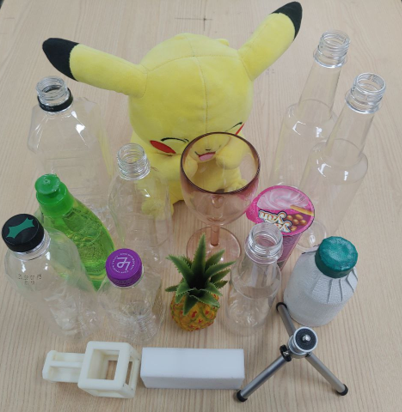}
    \caption{Objects used in the real-world experiments.}
    \vspace{-26pt}
    \label{fig:real_world_objects}
\end{wrapfigure}
In this section, we demonstrate that DreamGrasp provides sufficiently accurate recognition results for downstream use by executing two manipulation tasks in a cluttered shelf environment. We place various real-world objects including non-symmetric objects and transparent objects shown in Figure~\ref{fig:real_world_objects} on a shelf and perform object recognition.

\textbf{Instance Masks and Text Prompts.}
We use the state-of-the-art segmentation model SAM~\cite{kirillov2023segany} to obtain instance masks from real-world images. For text prompts, we generate object descriptions using ChatGPT~\cite{openai2024gpt4o} by querying the identity of each segmented object based on the raw image.

\textbf{Grasp Planning for Baselines.}
We evaluate CGC and Zero123* baselines using the same grasp planner as DreamGrasp. Note that DreamGrasp performs collision checking between surrounding objects and the grasped object along the retrieval trajectory. However, since baselines do not include a refinement stage, they often produce noisy instance predictions (as shown on the left of Figure~\ref{fig:segmentation_results}), leading to frequent false-positive collisions between the target and non-target objects. Therefore, we omit collision checking between the  target and non-target objects in the baseline grasp planning.

\subsubsection{Sequential Declutter}
\textbf{Scenarios.}
We conduct five trials for scenes containing two and four objects, respectively. A trial is considered successful if all objects in the scene are successfully grasped and removed without any collision; failure is recorded otherwise.

\begin{wraptable}{r}{4.7 cm}
\centering
\vspace{-29pt}
\caption{\label{table:grasping} Real-world single-shot recognition collsion-free declutter success rates on a shelf.}
\vspace{4pt}
\begin{tabular}{lccc}
\multicolumn{1}{c}{} &\multicolumn{2}{c}{Object \#} \\
\cline{2-3}
{\bf METHOD}  & 2 & 4 \\
\hline 
CGC & 1/5 & 0/5 & \\
Zero123* & 4/5 & 0/5  \\
DreamGrasp & {\bf 5/5} & {\bf 5/5} \\
\hline
\end{tabular}
\vspace{-10pt}
\end{wraptable}

\textbf{Results.}
Table~\ref{table:grasping} presents the experimental results. When only partial views are provided, CGC fails to reconstruct accurate geometry, resulting in low success rates. Zero123* successfully predicts reliable scene geometry and often succeeds in grasping. However, it cannot detect collisions with surrounding objects during retrieval. While this limitation has little effect in simpler two-object scenarios, it leads to frequent failures in four-object scenes due to collisions with nearby objects. In contrast, DreamGrasp achieves high success rates across all scenarios by enabling accurate instance-wise geometry estimation and effective collision-aware planning. A successful example of decluttering is shown in Figure~\ref{fig:clear_clutter}.

\subsubsection{Target Retrieval}

\textbf{Scenarios.}
We define success in the target retrieval task as successfully rearranging the obstacles and retrieving a designated target object without collisions from a shelf containing four objects; failure is recorded otherwise. Since prior baselines do not reliably detect collisions between the target and surrounding obstacles, we report results only for manipulations performed using DreamGrasp.

\begin{figure}[!t]
    \centering
    \vspace{-20pt}
    \includegraphics[width=\linewidth]{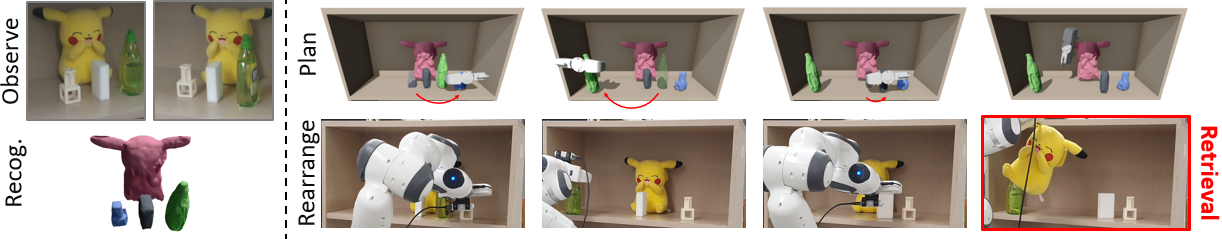}
    \vspace{-15pt}
    \caption{Example of real-world target retrieval on a shelf using DreamGrasp.}
    \vspace{-10pt}
    \label{fig:real_target_retrieval}
\end{figure}

\textbf{Results.}
We achieved an 80\% success rate over five trials. Figure~\ref{fig:real_target_retrieval} illustrates a successful retrieval trajectory. In this scenario, three obstacles are placed in front of a stuffed toy located at the back of the shelf. DreamGrasp accurately estimated the 3D shape and position of both the target object and surrounding obstacles, enabling a successful rearrangement and collision-free retrieval of the target.

\section{Conclusion}
We presented DreamGrasp, a zero-shot 3D reconstruction and instance recognition framework that operates from a few partial-view RGB images without requiring depth input or 3D training datasets. By combining the generative prior of large-scale 2D diffusion models with the structured 3D representation of Gaussian Splatting, DreamGrasp reconstructs accurate instance-wise geometry in cluttered, occluded scenes. The framework integrates coarse reconstruction via view-conditioned score distillation, instance segmentation through contrastive learning with a surface-invariant feature regularizer (SIFR), and instance-wise refinement guided by text prompts. Extensive experiments show that DreamGrasp not only achieves reliable 3D reconstruction at both scene and instance levels, but also enables effective robotic manipulation—demonstrating high success rates in real-world decluttering and target retrieval tasks using only a single recognition step.


\clearpage


\section{Limitations and Future Work}

While DreamGrasp demonstrates strong performance in instance-wise 3D reconstruction and downstream manipulation tasks, several limitations remain that suggest directions for future work.

One observed limitation is DreamGrasp’s difficulty in accurately reconstructing deeply concave regions, such as the interior of bowls or cups. As shown in Figure~\ref{fig:recog_results}, the top region of a cup often appears overfilled in the reconstructed mesh, resulting in inaccurate depth values inside the cavity. This arises from the fact that DreamGrasp relies solely on RGB information; since the reconstruction is guided by rendering outputs that appear plausible to a diffusion model, the system may produce geometry that merely \textit{looks} correct from the input views -- even if the actual 3D surface is incorrect. To address this, future work could incorporate estimated normal maps from novel views~\cite{long2024wonder3d} or integrate depth priors using fine-tuned depth foundation models~\cite{depthanything} to better constrain concave geometry.

Secondly, in the instance-wise refinement stage, DreamGrasp uses text -- conditioned score distillation to align each object’s rendered appearance with a given text prompt. However, when the guidance scale is too high, the model tends to overfit to the text condition and may generate an object that no longer resembles the original input scene. Careful hyperparameter tuning is required to balance fidelity to the input with faithfulness to the text. To alleviate this, future work could draw on techniques such as RealFusion~\cite{melas2023realfusion}, where a conditioning text is optimized to match the observed input image, potentially enabling more robust and scene-consistent refinement.

Lastly, we envision expanding DreamGrasp by distilling additional features -- beyond instance segmentation -- into the 3D Gaussians. These may include CLIP~\cite{radford2021learning}, DINO~\cite{caron2021emerging}, or other vision-language embeddings, enabling applications such as language-conditioned target grasping and object search. By integrating DreamGrasp with other large pre-trained models, we aim to further enhance its generalization and task -- level capabilities in open-world robotic settings.

\acknowledgments{This work was supported in part by IITP-MSIT grant RS-2021-II212068 (SNU AI Innovation Hub), RS-2022-II220480 (Training and Inference Methods for Goal Oriented AI Agents), RS-2024-00436680 (Collaborative Research Projects with Microsoft Research), KIAT grant P0020536 (HRD Program for Industrial Innovation), SRRC NRF grant RS-2023-00208052, 
SNU-AIIS, SNU-IPAI, SNU-IAMD, SNU BK21+ Program in Mechanical Engineering, SNU Institute for Engineering Research, and a KIAS Individual Grant (AP092701) via the Center for AI and Natural Sciences at Korea Institute for Advanced Study.}

\bibliography{references}  

\clearpage
\section*{Appendix}
\appendix


\section{Related Works}
\subsection{3D Recognition for Robotic Object Manipulation}

Previous studies on 3D instance shape recognition have primarily relied on depth images obtained from stereo RGB images. Depth information provides partial point clouds of object surfaces, which are commonly used to fit the pose of known objects~\cite{tan20176d} or to predict the shapes of unknown objects using neural networks~\cite{kim2022dsqnet, cai2022ove6d}. However, depth-based shape recognition suffers from real-world depth sensor noise, and reliable depth images cannot be obtained for transparent objects with non-Lambertian surfaces, such as glass.

To overcome these issues, some studies have attempted to recognize object shapes using only RGB images. Certain approaches train neural networks via supervised learning to predict depth images or complete imperfect depth images from RGB inputs~\cite{cao2022dgecn, sajjan2020clear, tang2021depthgrasp, fang2022transcg}, while others directly predict object shapes~\cite{kim2024t2sqnet, dai2023graspnerf}. These methods show strong performance in environments similar to their training settings. However, their performance is not guaranteed in out-of-distribution conditions—for example, with unseen camera poses or novel objects. While improving generalization requires diverse data, building large-scale 3D datasets is far more expensive than collecting 2D images, making it difficult to match the data scale of 2D tasks.

Another line of work for predicting 3D shape from RGB images without relying on 3D training data leverages differentiable 3D representation and rendering techniques, such as Neural Radiance Fields (NeRF)~\cite{mildenhall2021nerf}. These methods recognize the 3D shape of objects by optimizing 3D representations based on RGB images captured from all-around views of the scene, and then use these representations for grasping~\cite{ichnowski2021dex, kerr2022evo, rashid2023language, lee2023nfl}. Since they do not require separate training datasets, they are free from dataset construction costs. However, these methods require all-around views of the scene, and thus cannot guarantee reliable performance in environments with occlusions—such as shelves—where only partial observations are possible.

In contrast, our proposed method, DreamGrasp, is free from the limitations described above. DreamGrasp does not require 3D training data and benefits from the generalizability of diffusion models pretrained on large-scale 2D image datasets. Furthermore, it requires only a front partial view, making it suitable for scenarios with restricted visibility, such as shelf environments.

\subsection{3D Model Generation Guided by 2D Diffusion Models}

One approach to leveraging the rich knowledge of large-scale pretrained 2D diffusion models for 3D reconstruction is to optimize differentiable 3D representations(e.g., NeRF~\cite{mildenhall2021nerf}, NeuS~\cite{wang2021neus}, Gaussian Splatting~\cite{kerbl20233d}) such that their rendered images appear plausible to the diffusion model. Initial works~\cite{poole2022dreamfusion, wang2023score} have adopted this idea to generate 3D shapes from text prompts using text-conditioned diffusion models. Subsequent studies have extended this approach to front-view image inputs by training diffusion models to produce novel views, enabling single-image-based 3D generation~\cite{liu2023zero, qian2023magic123, tang2023dreamgaussian, liu2023one, liu2023syncdreamer, long2024wonder3d, tang2024mvdiffusion++}.

These methods mainly focus on generating high-fidelity 3D models of single, centered objects. Improvements have included estimating normal maps~\cite{long2024wonder3d} or enforcing view consistency across multiple novel views~\cite{tang2024mvdiffusion++}. However, most of these works target generative quality rather than recognition, and thus do not generalize well to multi-object scenes.

Meanwhile, a prior work, Part123~\cite{liu2024part123}, performs 3D reconstruction using an image-conditioned diffusion model and contrastive learning with mask images, similar to DreamGrasp. It combines single-object generation with part-level segmentation by using contrastive learning and masks generated by large models like SAM~\cite{kirillov2023segany} from rendered novel views. This approach is feasible because single-object generation is already sufficiently reliable. In contrast, DreamGrasp handles multi-object scenes, where novel-view renderings may be incomplete or inaccurate, making mask generation from rendered images unreliable. Therefore, instead of relying on SAM for novel-view mask generation, we propose a new regularizer that directly addresses the lack of mask information without relying on rendering quality.

\section{Details on Coarse Scene Reconstruction Stage}

\subsection{Detailed Preliminaries on Diffusion Models and Score Distillation Sampling}

\textbf{Diffusion Models.} 
Diffusion models~\cite{sohl2015deep, ho2020denoising} are a class of generative models that learn to model complex data distributions through a denoising process. The forward process, also called the diffusion process, gradually adds Gaussian noise to a clean data sample $I_0$ over a sequence of $T$ timesteps, resulting in a noised sample $I_t$ at each step. This process is typically defined as $I_t = \sqrt{\alpha_t}I_0 + \sqrt{1-\alpha_t}\epsilon$, where $\epsilon \sim \mathcal{N}(0, \textbf{I})$ is standard Gaussian noise and $\{\alpha_t\}_{t=1}^T$ is a predefined noise schedule controlling the amount of noise added at each timestep.

A denoising neural network $\epsilon_\theta(t, I_t)$ is trained to predict the noise $\epsilon$ given the noisy input $I_t$ and the timestep $t$. The model is trained to minimize the following denoising objective, 
\begin{equation}
    \mathcal{L}_{\text{DM}} = \mathbb{E}_{t\sim\text{Unif}\{1,\dots,T\}, \epsilon \sim \mathcal{N}(0, \textbf{I})}\left[w(t)||\epsilon_\theta(t, I_t) - \epsilon||^2 \right],
\end{equation}
where $w(t)$ is a time-dependent weighting function.

Recent extensions incorporate conditioning information $y$ (e.g., a text prompt or image feature) into the model, allowing for conditional generation by training $\epsilon_\theta(t, I_t, y)$. These works have widely used \emph{classifier-free guidance}~\cite{ho2022classifier} to improve conditional generation, which interpolates between the conditional and unconditional predictions to amplify the influence of the condition. Specifically, the guided prediction is $\hat{\epsilon}_\theta(t, I_t, y) = (1 + s)\epsilon_\theta(t, I_t, y) - s\epsilon_\theta(t, I_t)$, where $s$ is a guidance scale parameter that controls the strength of the condition. Larger values of $s$ lead to samples that more strongly reflect the condition $y$, at the potential cost of reduced diversity. 

\textbf{Score Distillation Sampling.}
Score Distillation Sampling (SDS)~\cite{poole2022dreamfusion} is a technique that enables optimization of 3D representation like NeRF or Gaussian splatting by leveraging the gradients of a pretrained diffusion model. Given the parameters of the 3D representation $\phi$, a differentiable renderer $\pi_X$ at camera pose $X$, a rendered image $\pi_X(\phi)$ and a target condition $y$ (such as a text prompt~\cite{poole2022dreamfusion} or an image~\cite{liu2023zero}), SDS computes gradients from $\hat\epsilon_\theta$ to guide $\phi$ toward satisfying the condition $y$. Specifically, the gradient is defined as:
\begin{equation}
    \nabla_\phi\mathcal{L}_{\text{SDS}} = \mathbb{E}_{t\sim\text{Unif}\{1,\dots,T\}, \epsilon \sim \mathcal{N}(0, \textbf{I}), X\sim\mathcal{X}}\left[w(t)(\hat{\epsilon}_\theta(t, \pi_{X}(\mathcal{\phi})_t, y) - \epsilon)\nabla_\phi \pi_{X}(\mathcal{\phi})\right],
\end{equation}
where $\pi_{X}(\mathcal{\phi})_t = \sqrt{\alpha_t}\pi_{X}(\mathcal{\phi}) + \sqrt{1 - \alpha_t} \epsilon$ is a noised version of the input $\pi_X(\phi)$. The random camera pose $X$ is sampled from $\mathcal{X}$, which is a uniform distribution of camera poses located on a hemisphere centered at the workspace center, each oriented to look toward the center of the workspace. Minimizing this loss allows the rendered image $\pi_X(\phi)$ to align with the distribution learned by the diffusion model, effectively guiding the 3D structure using the knowledge embedded in the 2D model. To effectively guide 3D model generation, prior works have shown that using the classifier-free guided prediction $\hat{\epsilon}_\theta(t, I_t, y)$ with high guidance scale $s$, instead of the raw denoiser output $\epsilon_\theta(t, I_t, y)$ leads to improved performance.

\subsection{Zero123 Leveraging Multi-View Input Images}
While the training strategy of Zero123 generates a high-quality 3D model consistent with \textit{a single front-view image}, it is not sufficient from a recognition standpoint due to the inherent \textit{scale ambiguity} -- that is, it cannot distinguish between a large object far from the camera and a small object close to it -- in single-image 3D reconstruction. To address this limitation, we leverage at least two partial-view RGB images $\{I_{j}\}_{j=1}^{N_I}$ with camera poses $\{X_{j}\}_{j=1}^{N_I}$ as input and train $\mathcal{G}$, where $N_I$ is the number of the observed partial-view images.

Let's denote the denoiser of novel-view diffusion model $\epsilon_\text{z}(t, I^{X}_t, I_\text{front}, X)$ where $I_t^X$ is the image at diffusion step $t$ from viewpoint $X$, $I_\text{front}$ is a front view image and $X$ is a relative camera pose. Additionally, note that since there are now multiple reference camera poses $\{X_{j}\}_{j=1}^{N_I}$, we define the notation as follows for clarity: $X$ denotes a randomly sampled novel view camera pose and $\Delta X_j$ refers to the relative pose of $X$ with respect to a reference camera pose $X_{j}$. Then, 
\begin{align}
    &\mathcal{L}_{\text{render}} = \sum_{j=1}^{N_I}\lVert\pi_{X_j}(\mathcal{G}) - I_j\rVert^2, \\
    &\nabla\mathcal{L}_{\text{Zero123}} = \sum_{j=1}^{N_I}\mathbb{E}_{\epsilon \sim \mathcal{N}(0, \textbf{I}), X\sim\mathcal{X}}\big[w(t)(\hat{\epsilon}_\text{z}(t, \pi_{X}(\mathcal{G})_t, I_j, \Delta X_j) - \epsilon)\nabla \pi_{X}(\mathcal{G})\big],
\end{align}
where $\pi_{X}(\mathcal{G})_t$ is a noised image of $\pi_{X}(\mathcal{G})$ with time step $t$ and $\hat{\epsilon}_\text{z}$ is corresponding classifier-free guidance of $\epsilon_\text{z}$. We use guidance scale of $s=3$ for classifier-free guidance.
To further improve the reconstruction quality, instead of randomly sampling 
$t$ from a uniform distribution, we adopt a predefined scheduling strategy~\cite{huang2023dreamtime} where $t$ deterministically varies with training epochs.

\subsection{Contrastive Learning Loss}



The feature of a pixel $\textbf{f}$ can be rendered by $\alpha$-blending, $\textbf{f} = \sum_{i \in \mathcal{N}} f_i \alpha_i \prod_{j=1}^{i-1}(1-\alpha_j)$. We also $L_2$-normalize $\text{f}$ so that $\textbf{f} \in S^{D_f-1}$ after rendering. With each mask $M_j$ corresponding to the camera pose $X_j$, we first render feature image from $X_j$ and extract the set of rendered feature vectors corresponding to the pixels belonging to the $k$-th object instance, denoted as $\textbf{f}_{j,k}$. Also, let $\overline{\textbf{f}}_{j,k}$ be the mean of the features in $\textbf{f}_{j,k}$. Then the contrastive learning loss is given as
\begin{align}
    \mathcal{L}_\text{CC} = \sum_{j=1}^{N_I}\sum_{k=1}^{N_s}\sum_{\textbf{f}\in \textbf{f}_{j,k}}-\log\frac{\exp(\textbf{f}\cdot\overline{\textbf{f}}_{j,k}/\psi^k)}{\sum_{l=1}^{N_s}\exp(\textbf{f}\cdot\overline{\textbf{f}}_{j,l}/\psi^l)},
\end{align}
where $N_s$ is the number of instances and $\psi^k$ is a temperature of the $k$-th object mask introduced in~\cite{silva2024contrastive}.

\subsection{Training}

The resulting training loss is given as
\begin{equation}
    \mathcal{L}_\text{coarse} = \lambda_\text{render}\mathcal{L}_\text{render} + \lambda_\text{Zero123}\mathcal{L}_\text{Zero123} + \lambda_\text{CC}\mathcal{L}_\text{CC} + \lambda_\text{SIFR}\mathcal{L}_\text{SIFR},
\end{equation}
where $\mathcal{L}_\text{reg}$ is other additive loss terms used in 3D asset generation frameworks and $\lambda_\text{render}$, $\lambda_\text{Zero123}$, $\lambda_\text{CC}$, $\lambda_\text{SIFR}$ are weights for each loss. 

We train the coarse stage for 1000 iterations using the loss functions described above. The rendering loss weight $\lambda_\text{render}$ is set to 500 for the first 100 iterations, linearly increased to 1000 by iteration 400, and then kept fixed at 1000 thereafter. The Zero123 supervision weight $\lambda_\text{Zero123}$ is fixed at 0.1 throughout training. $\mathcal{L}_\text{CC}$ and $\mathcal{L}_\text{SIFR}$ are not applied during the first 500 iterations, allowing the object shapes to reach a reasonable initial geometry before learning instance features. After 500 iterations, $\lambda_\text{CC}$ is set to 0.001 and $\lambda_\text{SIFR}$ to 0.0001. Note that to ensure the instance features influence only segmentation and not geometry, the gradients of $\mathcal{L}_\text{CC}$ and $\mathcal{L}_\text{SIFR}$ are restricted to flow only through the Gaussian features $f_i$.
\section{Details on Instance-wise Refining Stage}

\subsection{Solving Mask Correspondence}
Note that now we can resolve the correspondence problem across masks by comparing each mean rendered feature $\overline{\textbf{f}}_{j,l}$ of $l$-th object at other masks $M_j$ with the representative feature $\overline{f}_k$. Specifically, we solve bipartite matching problem $\hat{s}_j = \text{argmin}_{s \in \mathcal{S}_{N_s}}\sum_{k=1}^{N_s} \lVert\overline{f}_k  - \overline{\textbf{f}}_{j,s(k)}\rVert_2$, where $\mathcal{S}_{N_s}$ denotes the set of all permutations over $N_s$ elements. By applying the optimal alignment $\hat{s}_j$ to each mask image $M_j$, we ensure that the $k$-th object in every mask image corresponds to the $k$-th object in the representative mask image.

\subsection{Instance-wise Score Distillation Sampling with Text-based Diffusion Model}

We obtain text prompts for each instance from a single representative instance mask, and use these text prompts to refine the corresponding instances.
Let $\textbf{t}^k$ denote the text description corresponding to the $k$-th object. The rendered image $\pi_{X^k}(\mathcal{G}^k)$, obtained from the Gaussians $\mathcal{G}^k$ of the $k$-th object from the camera pose $X^k$, should be consistent with the text prompt $\textbf{t}^k$. Therefore, we utilize a text-conditioned diffusion model $\epsilon_\text{s}(t, I_t, \textbf{t})$ to obtain instance-wise SDS gradients as follows;

\begin{equation}
    \nabla\mathcal{L}_{\text{instance\_SDS}} = \sum_{k=1}^{N_s}\mathbb{E}_{\epsilon \sim \mathcal{N}(0, \textbf{I}), X^k\sim\mathcal{X}^k}\left[w(t)(\hat{\epsilon}_\text{s}(t, \pi_{X^k}(\mathcal{G}^k)_t, \textbf{t}^k) - \epsilon)\nabla \pi_{X^k}(\mathcal{G}^k)\right],
\end{equation}
where $\pi_{X^k}(\mathcal{G}^k)_t$ is a noised rendered image of $\pi_{X^k}(\mathcal{G}^k)$ with time step $t$ and $\hat{\epsilon}_\text{s}$ is corresponding classifier-free guidance of $\epsilon_\text{s}$. The instance-wise camera pose $X^k$ is sampled from $\mathcal{X}^k$, which is a uniform distribution of camera poses placed on a hemisphere centered at the center of the bounding box of each instance $\mathcal{G}^k$, with a radius large enough to ensure the entire bounding box fits within the image. Additionally, as in the Zero123 training in the coarse stage, we adopt a predefined schedule for $t$ instead of sampling it randomly. In practice, we use Stable Diffusion~\cite{rombach2021highresolution} as the text-conditioned denoiser $\epsilon_\text{s}$.

We use a guidance scale of $s = 10$ for classifier-free guidance. Note that this value is significantly lower than the original setting of $100$ proposed in DreamFusion~\cite{poole2022dreamfusion}. We observed that using the default guidance scale often causes the model to generate objects that align with the text prompt while ignoring the input RGB scene, thus breaking consistency. To balance the influence of the RGB rendering loss and Zero123 supervision, we empirically reduce the guidance scale to $10$.

\subsection{Details on Outlier Removal}
Let $\mu^k$ denote the set of mean values of $\mathcal{G}^k$. We apply the DBSCAN algorithm~\cite{ester1996density} with an epsilon value of $0.04$ to cluster $\mu^k$, and remove all clusters except the largest one. We perform this removal every 500 iterations after the refinement stage begins. Note that since this outlier removal step is performed periodically during training, even if valid geometry is mistakenly removed instead of noise, it can be recovered through continued optimization.

\subsection{Training}

We combine the proposed instance-wise refining loss with the Zero123 loss and the scene-level RGB reconstruction loss to form the final refinement objective;
\begin{equation}
    \mathcal{L}_\text{refine} = \lambda_\text{render}\mathcal{L}_\text{render} + \lambda_\text{Zero123}\mathcal{L}_\text{Zero123} + \lambda_\text{instance\_render}\mathcal{L}_\text{instance\_render} + \lambda_\text{instance\_SDS}\mathcal{L}_\text{instance\_SDS},
\end{equation}
where $\lambda_\text{render}$, $\lambda_\text{Zero123}$, $\lambda_\text{instance\_render}$, $\lambda_\text{{instance\_SDS}}$ are weights for each loss. The refining stage is run for 2000 iterations using the specified losses. We use the same weights as in the coarse stage for $\lambda_\text{render}$ and $\lambda_\text{Zero123}$, set to 1000 and 0.1, respectively. The weight for $\lambda_\text{instance\_render}$ is set to 1000. For $\lambda_\text{instance\_SDS}$, we observed that assigning a large weight early in training helps effectively suppress surrounding noise. However, maintaining a high weight throughout training disrupts the balance with other loss terms and leads to generations that match the text prompt but are inconsistent with the input RGB image. Therefore, we set $\lambda_\text{instance\_SDS}$ to 0.1 for the first 100 iterations, and reduce it to 0.01 thereafter.
\section{Details on Object Manipulation with DreamGrasp}

\subsection{Details on Grasp Pose Generation}
\label{appendix:grasp_pose}
To generate collision-free grasp poses for each instance, a sufficient number of valid candidates must remain after collision checking. Therefore, we generate a diverse set of grasp poses per object. To this end, we first sample all-around camera poses from a hemisphere centered on the workspace, render depth images from each pose, and use FC-GQ-CNN~\cite{satish2019policy}—a depth-based grasp pose predictor—to generate candidate grasps.

Note that FC-GQ-CNN was originally designed to predict top-down grasps for tabletop objects. To minimize domain shift, we adapt each instance’s depth image to resemble the tabletop scenes used during FC-GQ-CNN training. Specifically, we first reconstruct a mesh from instance $\mathcal{G}^k$ using TSDF fusion. Then, during depth rendering, we reposition the camera—while preserving its orientation—such that the farthest vertex of the mesh lies at a distance $d_{\text{far}}$, which corresponds to the camera-to-table distance used in FC-GQ-CNN’s training dataset. All background pixels (i.e., those not occupied by the object) are then filled with depth $d_{\text{far}}$. As a result, the rendered depth image resembles a scene in which the object is placed on a tabletop at distance $d_{\text{far}}$, enabling FC-GQ-CNN to generate reliable grasp poses. 

FC-GQ-CNN outputs a grasp quality score between $0$ and $1$ for each generated grasp pose. We retain only those poses with a score greater than or equal to $0.9$ for further use.

\subsection{Details on Collision Detection}
\label{appendix:collision}
For each candidate grasp pose, we verify whether the gripper collides with any non-target objects or the shelf during both the approach and the retrieval phases. We consider only a straight-line trajectory when the gripper approaches the object. Similarly, the retrieval phase is defined as a reversal of this approach trajectory, executed after the object has been grasped. We assume that the geometry of the shelf on which the objects are placed is known in advance, and the mesh model of the shelf is provided. 

To efficiently check for collisions between the gripper and other objects, we pre-process the gripper mesh using convex decomposition. All other objects are first reconstructed into meshes via TSDF fusion and then approximated by their convex hulls. For each discretized via-pose along the approach trajectory, we spawn the convex-decomposed gripper mesh and check for intersections with the shelf mesh and the convex hulls of all non-target objects. This allows us to detect collisions during both the approach and post-grasp holding phases.

During the retrieval phase, we treat the gripper and the grasped target object as a single rigid body. Specifically, we attach the convex hull of the target object to the gripper mesh at the grasp pose. For each via-pose along the reversed approach trajectory, we spawn this composite mesh and check for collisions with the convex hulls of surrounding non-target objects.

\subsection{Details on Sequential Declutter}
We leverage the instance-wise geometry provided by DreamGrasp to plan the entire decluttering sequence in a single step using only the initial recognition result. The procedure is as follows. First, we generate grasp poses for all recognized instances. Using the current scene configuration, we perform collision checking and retain only those grasp poses with collision-free approach trajectories. For each of the remaining candidates, we solve inverse kinematics, and randomly select one from the set of feasible grasp trajectories. Since each selected grasp pose is associated with a specific instance, we update the scene geometry by removing the corresponding object. We then repeat collision checking for the remaining grasp poses. This process is iteratively repeated until grasp trajectories have been planned for all objects in the scene.

\subsection{Details on Target Retrieval}

Let $\mathcal{P}$ denote the set of sampled grasping trajectories for a target object, which can be obtained using the grasp pose sampling and trajectory generation procedures described in Section~\ref{appendix:grasp_pose} and~\ref{appendix:collision}, and for which a valid inverse kinematics solution exists. We assume that the target object is initially ungraspable in the scene; that is, every trajectory $P \in \mathcal{P}$ results in a collision with the obstacles or the shelf. Additionally, we assume that when all obstacles are removed, there exists at least one trajectory $P \in \mathcal{P}$ that does not collide with the shelf and admits a valid inverse kinematics solution, enabling successful retrieval.

We define a collision number function $\mathcal{C}(P)$ as the number of obstacles and shelf regions that collide with a given trajectory $P$. Then, the graspability of the target object is defined as $\max_{P\in\mathcal{P}} -\mathcal{C}(P)$. The goal of object rearrangement is to maximize the target object's graspability via an appropriate sequence of pick-and-place actions $\{A_1, \dots, A_n\}$;
\begin{equation}
    \max_{\{A_1,\dots A_n\}}\max_{P\in\mathcal{P}}-\mathcal{C}(P).
\end{equation}

To solve this optimization problem, we adopt a sampling-based model predictive control (MPC) approach similar to~\cite{kim2023leveraging}. Since we can accurately predict the dynamics of obstacle pick-and-place actions, no learning-based components are used. Sampling of pick-and-place trajectories is performed only for those that are collision-free throughout the motion and admit valid inverse kinematics solutions, using the trajectory generation and collision-checking strategies described earlier.

\section{Details on Experiments and Additional Results}

\begin{figure}[!t]
    \centering
    \vspace{-20pt}
    \includegraphics[width=\linewidth]{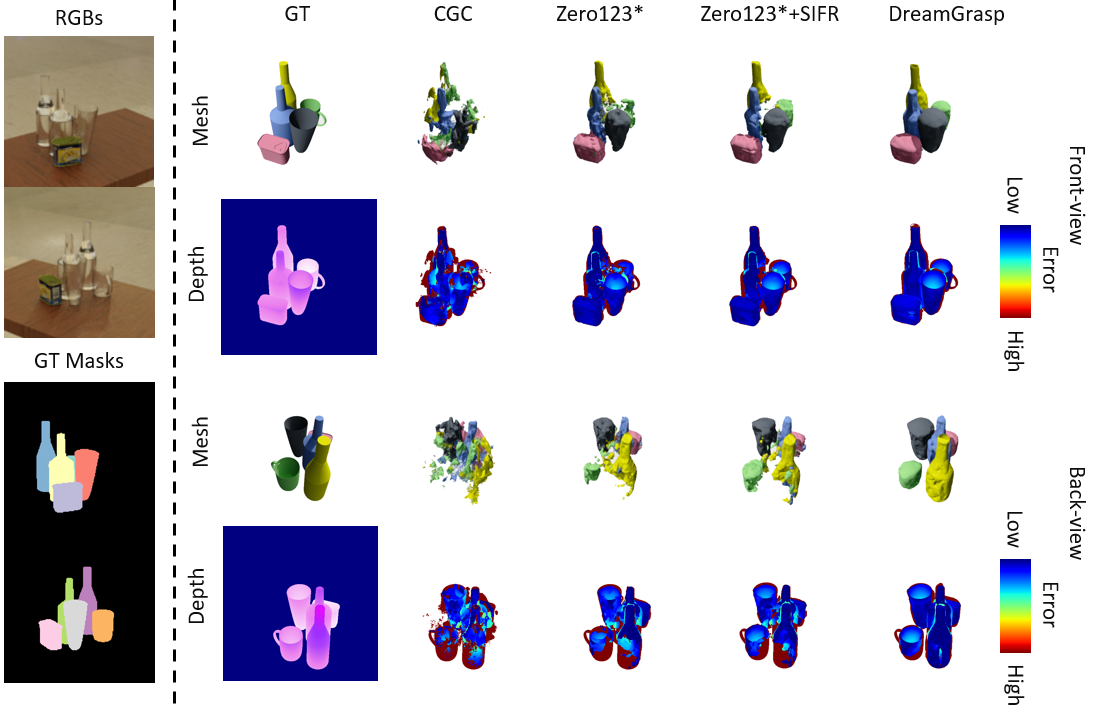}
    \vspace{-15pt}
    \caption{Extended Recognition results from two partial RGB images.}
    \vspace{-20pt}
    \label{fig:extended_recog_results}
\end{figure}

\subsection{Details on Experimental Settings}
For the recognition experiments, we sampled objects only from those in the YCB dataset that provide textured mesh files. From the TablewareNet dataset, we sampled objects from the following classes: wineglass, nottle, beerbottle, handlesscup, and mug. We excluded classes such as bowl and dish due to their large size, which often left insufficient space to place additional objects in the scene.

To evaluate depth accuracy and mask IoU, we rendered novel views from 80 camera poses uniformly sampled on a sphere of radius 1.5m centered at the workspace. Specifically, the camera positions were sampled by dividing the elevation angle from 0° to 60° into 4 steps and the azimuth angle from 0° to 360° into 20 steps, resulting in a total of 80 camera poses.

We used a 7-DOF Franka Emika Panda with a gripper and a RealSense D435 camera for the real-world experiments.

\subsection{Exteneded and Additional Recognition Experiments Results}

Table~\ref{table:recognition_extended} provides a detailed breakdown of recognition performance by the number of objects, including results for an additional baseline. Figure~\ref{fig:extended_recog_results} shows detailed visualization of recognition results.

\textbf{Zero123* + SIFR.}
CGC includes a spatial-similarity regularizer that enforces similar features among spatially close Gaussians and dissimilarity across distant ones. In contrast, DreamGrasp introduces the Surface-Invariant Feature Regularizer (SIFR), which encourages feature consistency specifically along the object surface. To isolate the impact of the regularizer, we replace CGC’s spatial-similarity regularizer in Zero123* with SIFR and compare the results.

\begin{table*}[!h]
\vspace{-15pt}
\centering
\caption{Extened Recognition Performance Table.}
\label{table:recognition_extended}
\begin{tabular}{lcc|cccccccc}

\multicolumn{3}{c}{}
&\multicolumn{3}{c}{Depth Acc. (Scene)}
&\multicolumn{1}{c}{}
&\multicolumn{3}{c}{Depth Acc. (Instance)}\\
\cline{4-6}
\cline{8-10}
\\
\multicolumn{1}{c}{\bf METHOD}
&\multicolumn{1}{c}{\bf Object \#}
&\multicolumn{1}{c}{}
&\multicolumn{1}{c}{$\delta_{0.05}$}
&\multicolumn{1}{c}{$\delta_{0.10}$}
&\multicolumn{1}{c}{$\delta_{0.20}$}
&\multicolumn{1}{c}{}
&\multicolumn{1}{c}{$\delta_{0.05}$}
&\multicolumn{1}{c}{$\delta_{0.10}$}
&\multicolumn{1}{c}{$\delta_{0.20}$}
&\multicolumn{1}{c}{Mask IOU}
\\

\hline 
\multirow{4}{*}{CGC}
&2&& 0.574 & 0.613 & 0.618 && 0.543 & 0.569 & 0.570 & 0.563 \\
&3&& 0.519 & 0.564 & 0.573 && 0.471 & 0.497 & 0.499 & 0.487 \\
&4&& 0.507 & 0.558 & 0.572 && 0.437 & 0.463 & 0.466 & 0.449 \\
&5&& 0.509 & 0.565 & 0.580 && 0.416 & 0.442 & 0.445 & 0.427 \\
\hline
\multirow{4}{*}{Zero123*}
&2&& 0.788 & 0.822 & 0.824 && 0.750 & 0.775 & 0.775 & 0.782 \\
&3&& 0.761 & 0.801 & 0.804 && 0.688 & 0.716 & 0.717 & 0.725 \\
&4&& 0.733 & 0.777 & 0.783 && 0.642 & 0.671 & 0.672 & 0.683 \\
&5&& 0.707 & 0.755 & 0.762 && 0.584 & 0.612 & 0.614 & 0.628 \\
\hline
\multirow{4}{*}{Zero123*+SIFR}
&2&& 0.762 & 0.798 & 0.800 && 0.716 & 0.742 & 0.743 & 0.748 \\
&3&& 0.750 & 0.792 & 0.796 && 0.681 & 0.710 & 0.711 & 0.711 \\
&4&& 0.729 & 0.774 & 0.780 && 0.649 & 0.678 & 0.679 & 0.675 \\
&5&& 0.707 & 0.755 & 0.763 && 0.607 & 0.635 & 0.636 & 0.631 \\
\hline
\multirow{4}{*}{DreamGrasp}
&2&& \textbf{0.812} & \textbf{0.850} & \textbf{0.852} && \textbf{0.802} & \textbf{0.831} & \textbf{0.832} & \textbf{0.821} \\
&3&& \textbf{0.791} & \textbf{0.838} & \textbf{0.840} && \textbf{0.758} & \textbf{0.788} & \textbf{0.788} & \textbf{0.771} \\
&4&& \textbf{0.770} & \textbf{0.820} & \textbf{0.824} && \textbf{0.734} & \textbf{0.767} & \textbf{0.768} & \textbf{0.742} \\
&5&& \textbf{0.747} & \textbf{0.800} & \textbf{0.806} && \textbf{0.703} & \textbf{0.735} & \textbf{0.736} & \textbf{0.701} \\
\hline

\end{tabular}
\vspace{-10pt}
\end{table*}

\begin{wrapfigure}{r}{0.3\textwidth}
    \centering
    \vspace{-5pt}
    \includegraphics[width=1\linewidth]{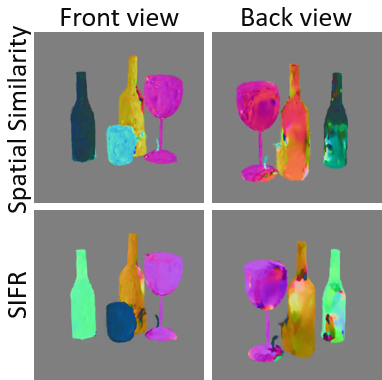}
    \caption{qualitative comparison of two regularizer: spatial-similiarity and SIFR.}
    \vspace{-20pt}
    \label{fig:ss_vs_sifr}
\end{wrapfigure}

Experimental results show that, in terms of instance-wise depth accuracy, Zero123* with the spatial-similarity regularizer outperforms the SIFR variant when the number of objects is small. However, as the number of objects increases, the performance of the SIFR variant surpasses that of the spatial-similarity regularizer. This trend can be explained as follows: spatial similarity regularization in Zero123* works well when objects are well separated but causes feature leakage between adjacent objects in cluttered scenes. In contrast, SIFR maintains feature consistency along object surfaces, making it more robust to densely packed scenes. Figure~\ref{fig:ss_vs_sifr} illustrates this effect, where spatial similarity causes feature spillover from a pink wineglass into a neighboring yellow bottle -- an issue largely avoided by SIFR.

\subsection{Domain Adaptation from Real-World to Zero123}
Note that Zero123 is trained to generate novel view images using a camera with fixed intrinsic parameters. For accurate geometry prediction, the camera parameters used in real-world settings must match those used during training of Zero123. However, real-world cameras often vary in their intrinsic parameters, and retraining Zero123 for each variation would be inefficient. Fortunately, the effect of changes in intrinsic parameters can be approximated by simple image cropping. Let the real-world camera’s intrinsic parameters be $(f_x, f_y, c_x, c_y)$ and the Zero123 camera’s intrinsic parameters be $(f_x', f_y', c_x', c_y')$. Given the real-world RGB image $I_{\text{real}} \in \mathbb{R}^{H\times W\times 3}$, the formula to transform it into the image $I_{\text{cropped}} \in \mathbb{R}^{H'\times W'\times 3}$, which is adjusted for Zero123, is as follows;
\begin{equation}
    I_{\text{cropped}}(i, j) = I_{\text{real}}\left(
    (i-c_y')\frac{f_y}{f_y'}+c_y, (j-c_x')\frac{f_x}{f_x'}+c_x
    \right).
\end{equation}
This is only feasible when converting from a camera with a wider field of view to one with a narrower field of view. Fortunately, the virtual camera used in Zero123 has a relatively narrow field of view (20°) compared to most real-world cameras, allowing us to address this mismatch by simply cropping real-world images.

\subsection{Additional Real-World Manipulation Experiments Results}
We have uploaded videos of the real-world experiments at the following link for reference.

https://youtu.be/BfCdhoNRJz0

\end{document}